\documentclass[11pt,a4paper]{article}
\usepackage[hyperref]{emnlp2018}
\usepackage{times}
\usepackage{latexsym}
\usepackage{CJK}
\usepackage{paralist}
\usepackage{graphicx}
\usepackage{enumitem}

\usepackage{url}

\aclfinalcopy 
\def\aclpaperid{1655} 

\title{BCWS: Bilingual Contextual Word Similarity}

\author{Ta-Chung Chi\quad Ching-Yen Shih\quad Yun-Nung Chen \\
  National Taiwan University, Taipei, Taiwan\\
  {\tt \{r06922028,r06943182\}@ntu.edu.tw\quad y.v.chen@ieee.org} 
  }

\date{}

\begin{document}
\maketitle
\begin{abstract}
This paper introduces the first dataset for evaluating English-Chinese Bilingual Contextual Word Similarity, namely BCWS\footnote{\url{https://github.com/MiuLab/BCWS}}.
The dataset consists of 2,091 English-Chinese word pairs with the corresponding sentential contexts and their similarity scores annotated by the human. 
Our annotated dataset has higher consistency compared to other similar datasets.
We establish several baselines for the bilingual embedding task to benchmark the experiments.
Modeling cross-lingual sense representations as provided in this dataset has the potential of moving artificial intelligence from monolingual understanding towards multilingual understanding.
\end{abstract}

\section{Introduction}
Distributed word representations have made a huge impact in the field of NLP by capturing semantics in the low-dimensional vectors, namely, the word embeddings~\cite{mikolov2013distributed}.
However, a word is usually represented by a single vector, ignoring the polymesy phenomenon in language.
To deal with this problem, \citet{reisinger2010multi} first proposed multi-prototype embeddings of a word and motivated a new research direction for sense embedding learning.

Following the pioneering work, a lot of work proposed to improve the quality of both word and sense embeddings.
Several datasets about word-level similarity were collected for intrinsically evaluating the embedding performance,
such as WS-353~\cite{finkelstein2001placing}, MEN~\cite{bruni2012distributional}, RW~\cite{luong2013better}, and MC-30~\cite{faruqui2016problems}.
However, there are few datasets available in terms of sense-level evaluation.
The first one is the Stanford contextual word similarity (SCWS) proposed by~\citet{HuangEtAl2012}.
Although this dataset alleviated the polysemy issue, it is a pure English dataset, 
and the inter-annotator consistency of this dataset is only about 0.52 in terms of Spearman's rank correlation, which upper bounds the performance the models can achieve.
Another is the recently proposed Word in Context (WiC) dataset~\cite{pilehvar2018wic}, which frames the sense disambiguation as a binary classification task and has a reasonable inter-rater agreement rate, but it is also a pure English dataset.

Recently, several works attempted to focus on learning cross-lingual embeddings in one space~\cite{adams2017automatic}.
A set of well-learned cross-lingual word embeddings can directly benefit several downstream tasks, such as unsupervised machine translation~\cite{lample2017unsupervised,artetxe2017unsupervised}. 
In addition, 
\citet{camacho2017semeval} proposed the cross-lingual semantic similarity dataset in Semeval2017, which measures the semantic similarity of word pairs within and across five languages: English, Farsi, German, Italian and Spanish.
Although this dataset has high inter-annotator agreements (consistently in the 0.9 ballpark), it cannot evaluate sense similarity due to the lack of word contexts.
Therefore, the semantic similarity evaluation on this dataset may not be precise enough. 

Nevertheless, none of the cross-lingual datasets considers multi-sense issues, where a word in one language may have multiple translations in another language according to its different meanings. 
Because learning word-level embeddings is inadequate, the concept about sense embeddings should also be extended to cross-lingual embeddings.
To deal with the above drawbacks of the prior datasets, we introduce a large and high-quality bilingual contextual word similarity (BCWS) dataset, which includes 2,091 English-Chinese word pairs with their sentential contexts and the human-labeled similarity scores for evaluating cross-lingual sense embeddings.
This is the first and only bilingual word similarity dataset with sentential contexts for evaluating cross-lingual sense similarity.
Note that our collected dataset can also be used as a cross-lingual word similarity data,
although it is designed for evaluating multi-sense embeddings.



\section{Dataset Construction}
\begin{figure}
  \centering
  \includegraphics[width=\linewidth]
  {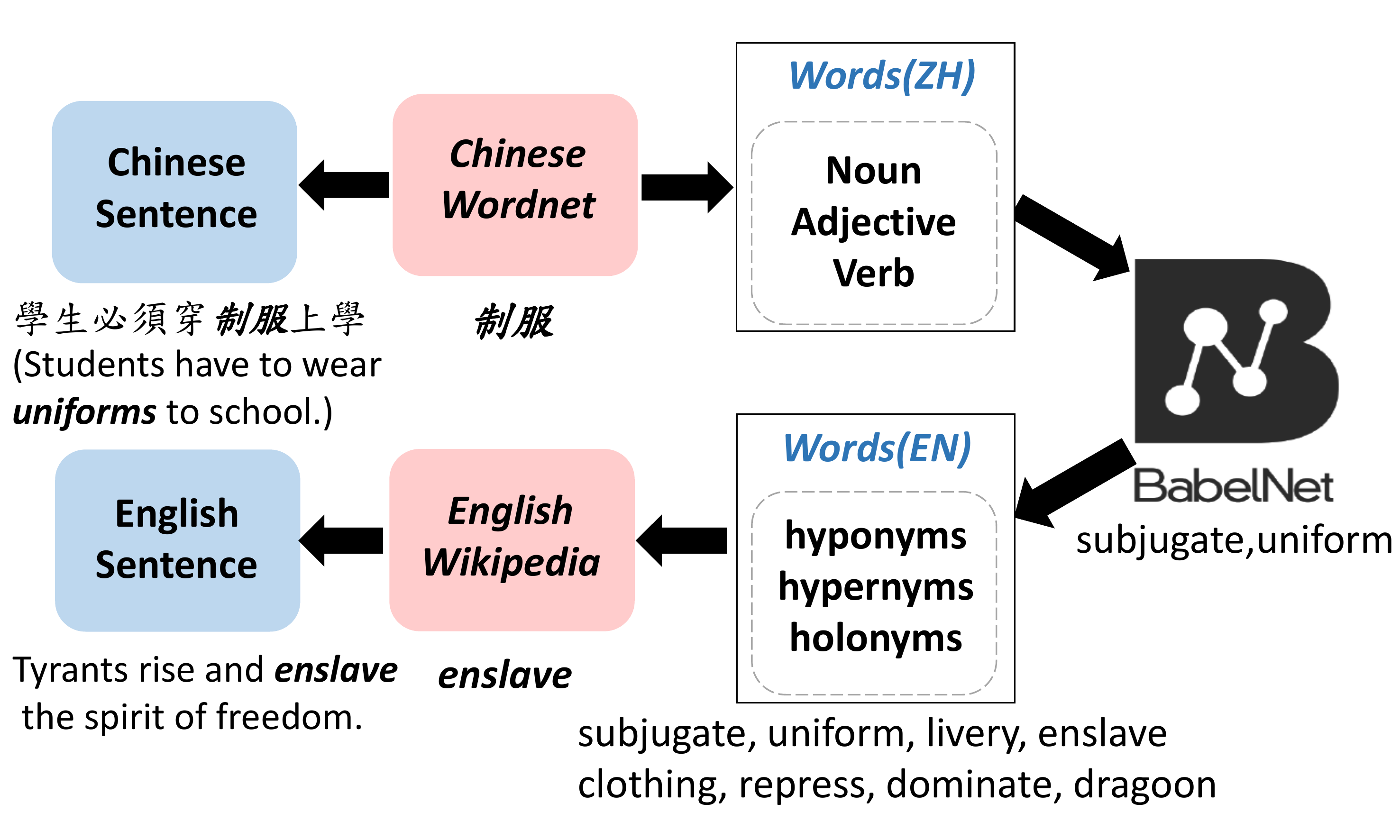}
  \vspace{-5mm}
  \caption{Illustration of the workflow.}
  \label{fig:workflow}
  \vspace{-3mm}
\end{figure}

To establish the bilingual contextual word similarity (BCWS) dataset, we collect the data by a five-step procedure as illustrated in Figure~\ref{fig:workflow}.

\subsection{Chinese Multi-Sense Word Extraction}
\label{ssec:ch_w}
First, we to extract the most frequent 10,000 Chinese words from Chinese Wikipedia dump.
Considering the common part-of-speech (PoS), we then select the words that are \emph{nouns}, \emph{adjective}, and \emph{verb} based on Chinese Wordnet~\cite{huang2010chinese}.
In order to test the sense-level representations, we remove words with only a single sense to ensure that the selected words are polysemous.
Also, the words with more than 20 senses are deleted, since those senses are too fine-grained and even hard for the human to disambiguate.
We denote the list of Chinese words $l_c$.

\begin{table*}[t]
\centering
\small
\begin{tabular}{|l|l|c|}
\hline
\bf English Sentence & \bf Chinese Sentence & \bf Score\\ 
\hline\hline
	Judges must give both sides an equal  
	&\begin{CJK}{UTF8}{bkai}我非常喜歡這個故事，它\textbf{$<$告訴$>$}我們一些重要的啟示。\end{CJK} & 7.00\\
	opportunity to \textbf{$<$state$>$} their cases.
    & (I like this story a lot, which \textbf{$<$tells$>$} us some important inspiration.) &\\
\hline
	It was of negligible \textbf{$<$importance$>$} prior 
	&\begin{CJK}{UTF8}{bkai}黃斑部病變的預防及早期治療是相當\textbf{$<$重要$>$}的。\end{CJK}  & 6.94\\
	 to 1990, with antiquated weapons and
    & (The prevention and early treatment of macular lesions is very & \\
     few members. & \textbf{$<$important$>$}.) &  \\
\hline
	Due to the San Andreas Fault bisecting  
	&\begin{CJK}{UTF8}{bkai}水果攤老闆似乎很意外真有人買這\textbf{$<$冷$>$}貨，露出「你真內行」\end{CJK} & 3.70\\
	 the hill, one side has \textbf{$<$cold$>$} water, the 
    &\begin{CJK}{UTF8}{bkai}的眼神與我聊了幾句。\end{CJK} (The owner of the fruit stall seemed surprised & \\
    other has hot. &  that someone bought this \textbf{$<$unpopular$>$} product, talking me few words &\\ & about ``you are such a pro''.) &\\
\hline
\end{tabular}
\vspace{-3mm}
\caption{\label{tb:bcws} Sentence pair examples and average annotated scores in BCWS.}
\vspace{-2mm}
\end{table*}

\subsection{English Candidate Word Extraction}
Second, the goal is to find an English counterpart for each Chinese word in $l_c$.
We utilize \emph{BabelNet}~\cite{navigli2010babelnet}, a free and open-sourced knowledge resource, to serve as our bilingual dictionary. 
Specifically, we first query the selected Chinese word using the free API call provided by Babelnet to retrieve all \textit{WordNet} senses\footnote{\emph{BabelNet} contains sense definitions from various resources such as Wordnet, Wikitionary, Wikidata, etc}.
For example, the Chinese word
``\begin{CJK}{UTF8}{bkai}制服\end{CJK}''
has two major meanings:
\begin{compactitem}
	\item \textit{uniform: a type of clothing worn by members of an organization}
	\item \textit{subjugate: force to submit or subdue}
\end{compactitem}
Hence, we can obtain two candidate English words, ``\textit{uniform}'' and ``\textit{subjugate}''.
Each word in $l_c$ retrieves its associated English candidate words, and then a dictionary $D$ is formed.

\subsection{Enriching Semantic Relationship}
Note that $D$ is merely a simple translation mapping between Chinese and English words.
It is desirable that we have more complicated and interesting relationships between bilingual word pairs.
Hence, for each English word in $D$, we find its \emph{hyponyms}, \emph{hypernyms}, \emph{holonyms} and \emph{attributes}, and add the additional words into $D$.
In our example, we may obtain \{\begin{CJK}{UTF8}{bkai}制服\end{CJK}: [uniform, subjugate, livery, clothing, repress, dominate, enslave, dragoon...]\}.
We sample 2 English words if the number of English candidate words is more than 5, 3 English words if more than 10, and 1 English word otherwise to form the final bilingual pair.
For example, a bilingual word pair (\begin{CJK}{UTF8}{bkai}制服\end{CJK}, enslave) can be formed accordingly.
After this step, we obtain 2,091 bilingual word pairs $P$.

\subsection{Adding Contextual Information}
Given the bilingual word pairs $P$, appropriate contexts should be found in order to form the full sentences for human judgment.
For each Chinese word, we randomly sample one example sentence in Chinese WordNet that matches the PoS tag we selected in \ref{ssec:ch_w}.
For each English word, we find all sentences containing the target word from the English Wikipedia dump.
We then sample one sentence where the target word is tagged as the matched PoS tag\footnote{We use the NLTK PoS tagger to obtain the tags.}.

\subsection{Human Labeling}

In order to associate a similarity measure with a collected bilingual word pair with their contexts, we recruit 11 human annotators for annotating the semantic scores.
To ensure the workers' proficiency, all recruited annotators are Chinese native speakers whose scores are at least 29 in the TOEFL reading section or 157 in the GRE verbal section.
All pairs will be scored by all 11 annotators in a random order.
To ensure consistency of labeling, the annotators are highly encouraged to look up a given dictionary, the English Oxford dictionary\footnote{\url{https://www.oxforddictionaries}}, due to its plentiful example sentences.
Note that they are asked not to rely solely on dictionary definitions but should consider the contextual information given in questions.

The annotators are asked to determine the sense similarity of these two target words based on their contexts in the sentences. 
Each question is given a score between 0.0 and 10.0 depending on how semantic related they are.
\begin{compactitem}
\item 0.0 indicates that the semantic meanings of the two target words are entirely different.
\item 10.0 indicates that the semantic meanings of two target words are entirely the same. 
\end{compactitem}
If a particular question is difficult to answer; for example, for the questions with terribly missing words that prevent them from understanding the meaning, the annotators can mark them with 0.0.
To ensure the same grading standard, the annotators are asked to finish all questions within 3 days, and we also retest some previously answered questions to make sure they receive similar scores.

\section{Data Analysis}

\begin{figure}
  \centering
  \includegraphics[width=\linewidth]{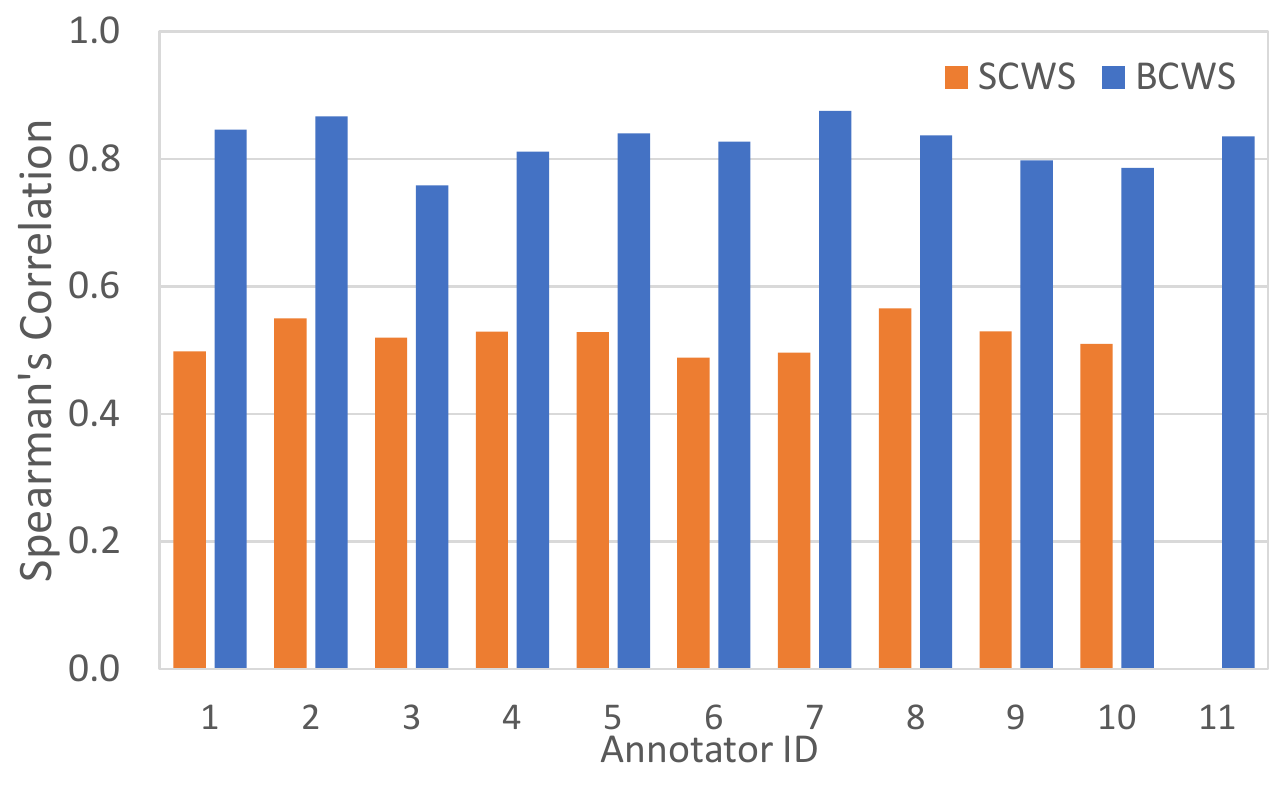}
  \vspace{-7mm}
  \caption{The distribution of the annotated Spearman's rank correlation computed by leave-one-out.}
  \vspace{-4mm}
  \label{fig:data}
\end{figure}

Our collected BCWS dataset includes 2,091 questions, each of which contains exactly one Chinese sentence and one English sentence.
Moreover, each sentence contains exactly one target word that is surrounded by $<$ and $>$ shown in Table~\ref{tb:bcws}.
After finishing labeling, the inter-annotator consistency is then calculated.
Specifically, we leave one annotator out and calculate the Spearman's rank correlation between the scores from the annotator who is left out and the average of the remaining annotators.
The average score can be viewed as the human performance, the upper bound of the embedding models.
The average agreement of BCWS is 0.83, while the agreement of previously similar dataset SCWS~\cite{HuangEtAl2012} is about 0.52.
The distribution of the correlation scores for two datasets is shown in Figure~\ref{fig:data}.
It can be found that our BCWS dataset has much higher consistency among annotators compared to SCWS, demonstrating the better quality for evaluating sense embeddings.

From the prior work on SCWS, the current state-of-the-art score is around 0.7, and most work cannot further improve the performance significantly, because they have already surpassed human-labeled performance on SCWS.
This observation is also pointed out by~\citet{pilehvar2018wic}.
Moreover, note that a merely 300-dimensional word-level skip-gram model can achieve a score of 0.65~\cite{bartunov2016breaking} on SCWS.
In contrast, our baseline word-level skip-gram model can only obtain a score of 0.49, indicating that our dataset provides a larger room of improvement for the follow-up work.

\section{Baseline Experiments}
We benchmark the experiments by presenting several baseline models about cross-lingual embeddings.
We assume that the sentence-level parallel corpus is available but without word-level alignments. 
The used parallel data is UM-corpus~\cite{tian2014corpus}, which contains 15,764,200 parallel sentences with 381,921,583 English words and 572,277,658 unsegmented Chinese words.
We exploit a widely-used tool \textit{jieba}\footnote{\url{https://github.com/fxsjy/jieba}} to perform Chinese word segmentation.
For those baseline models that train word-level embeddings, word similarity score can be obtained by calculating cosine similarity between two target words' embeddings.
Then the Spearman's rank correlation between human labeled scores and the cosine similarity scores is calculated to measure how well these two scores are correlated. 
We briefly introduce three baseline methods below and show all results in Table~\ref{baseline}. 

\vspace{-1mm}
\paragraph{Pretrained Word Vectors}
The na\"{i}ve baseline is to simply pretrain word embeddings of two languages. We use \texttt{word2vec}
to train word embeddings for Chinese and English parts of the UM-corpus~\cite{mikolov2013efficient}, where the default hyper-parameters settings are adopted. 
Obviously, this method has poor performance (1.16 for Spearman's rank), because it does not consider any interaction and alignment between the two languages.
In other words, these two sets of embeddings do not live in the same vector space.

\vspace{-1mm}
\paragraph{Bilingual Word Embeddings}
\citet{luong2015bilingual} proposed a bilingual word representation system which extends the skip-gram architecture to predict not only neighbor words in the same language, but also neighbor words in its bilingual counterpart.
It assumes that the system uses either the given ground truth word alignment or naive monotonic order alignment.
For a fair comparison, we experiment on the none word alignment version. 
This method directly trains cross-lingual word embeddings from scratch jointly.
We train 300-dimensional word vectors with 25 negative samples and leave other parameters as the default configuration. 
The achieved performance is 49.20 on Spearman's correlation, and the reason may be that the learned embeddings contain more noises during training due to the lack of word alignments, showing the difficulty of bridging the signal between two languages.

\vspace{-1mm}
\paragraph{Multilingual Word Embedding}
\citet{conneau2017word} proposed MUSE, an unsupervised method for mapping two sets of monolingual word embeddings into the same space via adversarial training.
It learns a transformation matrix \textit{W} which is nearly orthogonal and utilizes it to align two word embedding spaces. 
Adversarial training is applied to allow a randomly selected word to feed to the discriminator for determining which vector space the word belongs to.

This method requires two sets of pre-trained embeddings using \texttt{fasttext}~\cite{bojanowski2017enriching},
where we select 6,000 words with highest frequencies in each of Chinese and English parts of the UM-corpus and train 300-dimensional word vectors with the default settings. 
Then we perform adversarial matrix transformation for mapping the vectors into the same space and compute the correlation performance. 
Although the linguistic structure of English and Chinese are totally different, MUSE can still align two embedding spaces quite well, achieving 54.7 on Spearman's correlation.

\vspace{-1mm}
\paragraph{Bilingual Sense Embeddings}
\citet{chi2018cluse} proposed a first sense-level cross-lingual representation learning model with efficient sense induction, where several monolingual and bilingual modules are jointly optimized.
We train this model on the UM-corpus and achieve 58.5 on Spearman's correlation.

\begin{table}[t!]
\centering
\begin{tabular}{|l|r|}
\hline 
\bf Baseline Model & \bf Correlation \\ 
\hline\hline
 \citet{mikolov2013efficient} &  1.16  \\ 
\hline
 \citet{luong2015bilingual} &  49.20  \\ 
\hline
 \citet{conneau2017word} & 54.70  \\ 
 \hline
 \citet{chi2018cluse} & 58.80\\
\hline
 Human performance &  82.58  \\ 
\hline
\end{tabular}
\caption{\label{baseline} Result of current baselines. The reported numbers indicate Spearman's rank correlation $\rho\times 100$. }
\vspace{-3mm}
\end{table}

Although the result of sense embeddings is significant improved recently, all current results show the difficulty of learning bilingual sense embeddings. 
The proposed dataset still has a large room for improvement, offering a research  direction for future exploration.

\section{Conclusion}
We present the first dataset to provide evaluation for bilingual contextual word similarity.
Unlike the most word similarity datasets, this dataset measures word similarity given their sentential contexts in different languages.
Moreover, this dataset has high inter-annotator consistency, providing a large room for improvement towards human performance.
The new dataset has the potential of helping researchers explore a new direction of the cross-lingual word and sense embeddings and moving monolingual understanding towards multilingual understanding.


\bibliographystyle{acl_natbib_nourl}
\bibliography{emnlp2018}

\end{document}